\documentclass[openacc]{rsproca_new}%%%%where rsproca is the template name

\titlehead{Review}

\begin{document}

%%%% Article title to be placed here
\title{Introduction to the Special Issue on Symbolic Regression in the Physical Sciences}

\author{%%%% Author details
Deaglan J. Bartlett$^{1}$, Harry Desmond$^{2}$, Pedro G. Ferreira$^{1}$, Gabriel Kronberger$^{3}$}

%%%%%%%%% Insert author address here
\address{$^{1}$ Astrophysics, University of Oxford, Denys Wilkinson Building, Keble Road, Oxford OX1 3RH, UK\\
$^{2}$Institute of Cosmology \& Gravitation, University of Portsmouth, Dennis Sciama Building, Portsmouth, PO1 3FX, UK\\
$^{3}$Heuristic and Evolutionary Algorithms Laboratory, University of Applied Sciences Upper Austria, Softwarepark 11, 4232 Hagenberg, Austria
}

%%%% Subject entries to be placed here %%%%
% \subject{xxxxx, xxxxx, xxxx}

%%%% Keyword entries to be placed here %%%%

%%%% Insert corresponding author and its email address}
\corres{Deaglan J. Bartlett, Harry Desmond, Pedro G. Ferreira, Gabriel Kronberger\\
\email{deaglan.bartlett@physics.ox.ac.uk}, \email{harry.desmond@port.ac.uk}, \email{pedro.ferreira@physics.ox.ac.uk}, \email{Gabriel.Kronberger@fh-hagenberg.at}}

%%%% Abstract text to be placed here %%%%%%%%%%%%
\begin{abstract}

{Symbolic regression (SR) has emerged as a powerful method for uncovering interpretable mathematical relationships from data, offering a novel route to both scientific discovery and efficient empirical modelling. This article introduces the Special Issue on Symbolic Regression for the Physical Sciences, motivated by the Royal Society discussion meeting held in April 2025. The contributions collected here span applications from automated equation discovery and emergent-phenomena modelling to the construction of compact emulators for computationally expensive simulations.}

{The introductory review outlines the conceptual foundations of SR, contrasts it with conventional regression approaches, and surveys its main use cases in the physical sciences, including the derivation of effective theories, empirical functional forms and surrogate models. We summarise methodological considerations such as search-space design, operator selection, complexity control, feature selection, and integration with modern AI approaches. We also highlight ongoing challenges, including scalability, robustness to noise, overfitting and computational complexity. Finally we emphasise emerging directions, particularly the incorporation of symmetry constraints, asymptotic behaviour and other theoretical information. Taken together, the papers in this Special Issue illustrate the accelerating progress of SR and its growing relevance across the physical sciences.}

\end{abstract}

\begin{fmtext}
\end{fmtext}
\maketitle

%%%%%%%%%%%%%%%%%%%%%%%%%%%

% \rsbreak

%%%%%%%%%%%%%%%%%%%%%%%
% Gemini 2.5 Pro Preview
% Prompt:
%%%%%%%%%%%%%%%%%%%%%%%
% The attached file is an automatic transcription from an online meeting of four researchers discussing the state-of-the-art of symbolic regression in the physical sciences, and potential future directions of development. The transcription is not flawless and contains tags with timestamps to assign statements to persons. Ignore the tags completely. It is not relevant who stated what. Instead, focus solely on the content of the discussion.
% Summarise the content of the meeting in approximately 2500 words for an introductory article for a scientific journal for mathematics and natural sciences. The summary should focus on the main lines of the discussion and skip details or vague statements. The summary should state the facts and ignore that the source is a discussion of researchers. You are allowed to fill in information gaps using your background knowledge.
% Attachment: RS_intro_zoom_meeting_transciption.txt
%%%%%%%%%%%%%%%%

% Output generated by Gemini after editing:

\section{Introduction: Symbolic Regression and the Royal Society Meeting}

Symbolic Regression (SR) is emerging as a powerful computational methodology with significant potential to advance research across the physical sciences. This approach, distinct from conventional regression techniques that fit parameters to a predefined model structure, aims to uncover the underlying mathematical expressions that best describe observed data. By navigating the vast space of possible equations, SR offers a pathway both to create highly efficient predictive models and, more profoundly, to discover novel scientific laws or relationships from empirical evidence.

% GKR: added brief history of early AI techniques
Automatically finding equations that fit observational data has been a topic of interest in AI research since the early beginnings.
Early symbolic and heuristic systems for rule-based equation discovery include Lenat's Automatic Mathematician~\cite{Lenat1977}, BACON~\cite{Langley1981}, ABACUS~\cite{Falkenhainer1986}, or LAGRANGE~\cite{Dzeroski1995} for algebraic differential equations. Genetic programming (GP) is an evolutionary algorithm and was  introduced in the 90s and popularised by Koza~\cite{Koza1992}. Modern implementations of GP for symbolic regression are for example PySR~\cite{CranmerPySR} or PyOperon~\cite{BurlacuOperon}. Later, evolutionary and rule-based approaches were combined in systems such as AI Feynman~\cite{Udrescu2020}, or AI Descartes~\cite{Cornelio2023}. Systems relying on deep learning and reinforcement learning include EQL~\cite{EQL} and uDSR~\cite{Landajuela2022}. More recently, several different approaches based on end-to-end learning, transformers, and foundational models have been proposed.

At its core, symbolic regression endeavors to identify a functional form, $f$, such that $y=f(x_1,x_2,\ldots,x_n)$ optimally describes the relationship between a set of input variables $\{x_i\}$ and an output variable $y$, based on a given dataset. Unlike traditional regression methods where the structure of $f$ (e.g., linear, polynomial of a certain degree) is assumed a priori, SR algorithms, explore a diverse range of mathematical operators (e.g., arithmetic operations, trigonometric functions, exponentials, logarithms) and their combinations to construct candidate equations.

The primary output of SR is not just a set of parameters, but one or more explicit mathematical formulae. This inherent transparency is a key advantage, as the discovered equations are, in principle, {human-interpretable}. This interpretability allows scientists to gain insights into the underlying mechanisms of the system under study, validate the discovered models against existing theories, and potentially reveal previously unknown connections between variables. Furthermore, SR models, by capturing the intrinsic structure of the data, may exhibit superior generalisation capabilities, particularly when extrapolating beyond the range of the training data.

The growing interest in SR and equation discovery methods in particular in the physical sciences has led us to organize the discussion meeting on \textit{Symbolic Regression in the Physical Sciences} which was held in April 2025 at the Royal Society in London. {The gathering assembled a group of researchers at the interface of machine learning and the physical sciences, aiming to develop symbolic regression methods and use them to learn physical laws or simplified models directly from experimental or observational data. Applications across domains illustrated the versatility of SR, for example deriving constitutive laws for metallic materials under deformation, accelerating cosmological modelling and learning dark-matter halo profiles from kinematic data.}

{In this special issue we have collected research articles from the presenters at the discussion meeting. This introductory article outlines the main uses of SR, the evolving methodological frontier and the challenges that constitute the primary concerns of SR researchers moving forwards.}

% \section{The Essence of Symbolic Regression}

\section{Symbolic Regression for Scientific Discovery: The Quest for Physical Laws}

The most ambitious application of symbolic regression is in automated scientific discovery for extracting fundamental physical laws or novel descriptive equations directly from experimental or observational data.
% This harks back to the foundational moments of science, such as Kepler's derivation of the laws of planetary motion from Tycho Brahe's astronomical observations. Modern SR aims to systematise and accelerate this process using computational power.
The goal here is to find simple, yet accurate, mathematical formulae that capture the essence of  complex datasets. This aligns with the principle of parsimony, or Occam's Razor, which favors simpler explanations. SR algorithms often incorporate complexity penalties or simplicity priors to guide the search towards more concise and natural equations. Compact equations are more likely to represent true underlying relationships rather than mere overfitting. Challenges in this domain are numerous, including the presence of noise in data, the potential for observational biases, the vastness of the search space for equations, and the difficulty in objectively defining ``simplicity'' or ``interpretability''.

While the prospect of SR uncovering entirely new, fundamental laws of physics on par with general relativity or quantum mechanics is tempered by the understanding that many such foundational principles may have already been established, significant opportunities remain. The frontier for SR-driven discovery is particularly rich in areas characterised by {emergent phenomena}, where complex behaviours arise from the interactions of many simpler components, and where deriving macroscopic laws from microscopic descriptions is often intractable.
Examples of such situations in the physical sciences include:

\begin{itemize}
\item \textit{Astrophysics:} The Universe presents a plethora of complex systems, from plasma dynamics in stellar atmospheres and accretion disks to the formation of galaxies and the large-scale structure of the cosmic web. Highly nonlinear physics couples a vast range of scales, making such processes extremely difficult to simulate. SR can be applied to observational (e.g., light curves, spectra, cosmological surveys) or simulated data to find new relationships governing these phenomena, potentially leading to improved models for stellar evolution, galaxy dynamics, or the properties of exoplanetary atmospheres.
\item \textit{Condensed Matter Physics:} This field abounds with complex many-body systems exhibiting rich emergent behaviours, such as superconductivity, topological phases, or complex magnetic orderings. SR could help identify effective theories or phenomenological laws that describe these macroscopic properties from experimental measurements (e.g., transport, spectroscopic data) or from data generated by detailed microscopic simulations that are too complex to yield direct analytical insight.
\item \textit{Engineering:} SR can be employed to derive equations predicting performance metrics from design parameters in the design of complex systems (e.g., electrical motors, chemical reactors), to find constitutive models for materials, or to find optimal control strategies for dynamic processes.
\end{itemize}

In these domains, SR is not necessarily seeking to overturn established fundamental laws but rather to discover the specific mathematical forms that govern the emergent behavior within a particular context, often providing crucial links between microscopic physics and macroscopic observables.

\section{Symbolic Regression for Empirical Modeling}

Beyond the pursuit of fundamental laws, symbolic regression serves as a highly effective tool for fitting functions to data and generating empirical formulae. In many scientific and engineering contexts, the primary need is for a compact, accurate mathematical model that captures the observed relationships, even if it does not represent a physical law. SR can produce such empirical models that are often more insightful and easier to use than nonparametric or overparameterized (deep) models.

A particularly valuable attribute of SR in this context is its potential for robust extrapolation. Because SR aims to find the actual functional form of the relationship, if it succeeds in capturing even an approximation of the true underlying physics, the resulting model may extrapolate more reliably to unseen regions of the parameter space than models that merely interpolate based on local data patterns. This is crucial for making predictions under novel conditions or for understanding system behavior beyond the experimentally accessible range. However, like all extrapolation, this capability must be approached with caution, as a model that performs well on training data can still fail dramatically outside that domain if the discovered functional form is not truly representative of the underlying process. Simplicity priors and the minimum description length principle for SR play an important role in this context.

\section{Symbolic Regression for Emulation in Physical Systems}

An important example of creating fitting formulae is the development of emulators, also known as surrogate models. Many physical phenomena are described by complex, high-dimensional, and computationally intensive simulations derived from first-principles theories (e.g., quantum mechanics, fluid dynamics, general relativity). Running these simulations for extensive parameter sweeps, uncertainty quantification, or real-time control can be prohibitively expensive.

Symbolic regression offers a compelling solution by generating {fast and accurate mathematical approximations of these complex simulations}. These emulators, expressed as relatively simple equations, can reproduce the input-output behavior of the original simulation with high fidelity but at a fraction of the computational cost. For instance, SR can be used to derive analytical expressions for material properties that normally require demanding quantum mechanical calculations, or to create compact models of cosmology simulations, or to approximate solutions to partial differential equations governing fluid flow.

The benefit of SR-derived emulators extends beyond mere speed. Because the emulator is an explicit mathematical formula, it can provide insights into the sensitivity of the system's outputs to various input parameters, potentially revealing dominant physical effects or simplifying assumptions that might not be obvious from the full simulation. These effective descriptions of complex physics are invaluable for rapid design exploration, optimisation, and inverse problem solving, where numerous model evaluations are necessary. Unlike many alternative emulation tools such as neural networks, the outputs of SR are succinct mathematical expressions. These can be easily coded in any programming language and potentially be used even on resource-constrained hardware, such as embedded devices, e.g. for online control.

\section{Methodological Aspects}

The successful application of symbolic regression requires careful consideration of several methodological aspects. It is not a universal panacea, and its effectiveness is highly dependent on the nature of the problem and the way the search is conducted.

A crucial first step is {strategic problem selection}. SR is most likely to yield significant benefits when the goal is to obtain a potentially interpretable, explicit mathematical model, when there is a reasonable expectation that a concise underlying relationship exists. In situations where predictive accuracy is the sole concern and interpretability is secondary, or when a large number of potentially interrelated input variables have to be considered, other machine learning techniques such as deep neural networks might be more appropriate. These much larger models however require larger data volumes for training. While SR can work with smaller datasets, noisy or insufficient data may lead to spurious or overly complex equations and overfitting.

The inherent complexity of the physical system under study also impacts the application of SR. Highly complex systems with many interacting variables may not be amenable to description by simple, low-dimensional equations. In such cases, appropriate feature engineering, dimensionality reduction, or focusing on specific, well-constrained aspects of the system might be necessary. GP exhibits  implicit feature selection through evolutionary selective pressure, which allows to identify the most relevant input variables for low-dimensional equations from high-dimensional datasets automatically. However, the scalability of current SR algorithms can also be a limitation in situations where a large number of input variables must be captured in a large model, or when searching for large  mathematical expressions. Here new developments for the automatic identification of functions or hierarchical, modular structures are of interest to facilitate more compressed and compact descriptions.

Furthermore, the definition and constraining of the search space through the set of mathematical building blocks (variables, constants, operators) and the rules for combining them, is a critical determinant of SR's success. An overly restrictive search space might preclude the discovery of the more likely  equations, while an excessively large space can render the search computationally intractable or increase the risk of finding complex equations that overfit the data. The choice of basis functions should ideally be guided by domain knowledge about the system being modeled.

\section{The Evolving Frontier of Symbolic Regression}

The field of symbolic regression is dynamic, with ongoing research aimed at enhancing its power, scope, and reliability. Two particularly promising future directions involve the integration of prior scientific knowledge and the synergistic combination with recent AI technologies such as deep learning, reinforcement learning or large language models (LLMs). In this context, SR is a prime candidate where hybrid symbolic, sub-symbolic and neural AI systems may be especially helpful.

%\begin{itemize}
%\item {Incorporating Prior Scientific Knowledge and Constraints:}
A significant advancement in SR is the ability to move beyond purely data-driven approaches by incorporating existing scientific knowledge and physical constraints directly into the search process. This can improve the efficiency of the search and the physical plausibility of the discovered equations. Examples of such prior knowledge include:

\begin{itemize}
\item \textit{Symmetries:} Imposing known symmetries of the system (e.g., translational, rotational, parity) on the candidate equations.
\item \textit{Conservation laws:} Requiring that the discovered equations respect known conservation principles (e.g., conservation of energy, momentum, or mass).
\item \textit{Known asymptotic behaviors or boundary conditions:} Guiding the search towards functions that satisfy known limits or conditions.
\item \textit{Partial or incomplete theoretical models:} Using SR to discover missing terms or functional forms within an otherwise established theoretical framework.
\item \textit{Dimensional homogeneity:} Ensuring that all terms in an equation have consistent physical units.
\end{itemize}
Explicit incorporation of such constraints effectively reduces the search space and steers algorithms towards more meaningful and physically sound solutions.

%\item {Synergies with Large Language Models (LLMs):}
The combination of symbolic regression with foundation models such as large language models represents a nascent but highly promising avenue for future development. Foundation models, trained on vast corpora of text or code, possess remarkable capabilities in natural language understanding, generation, and even rudimentary reasoning. This opens up several possibilities for enhancing SR:
\begin{itemize}
    \item \textit{Hypothesis generation:} LLMs could analyse scientific literature or problem descriptions to suggest relevant physical variables, potential functional forms, or applicable physical principles, thereby helping to define and constrain the search space for SR.
    \item \textit{Interpretation and explanation:} LLMs could translate the often complex mathematical expressions discovered by SR into intuitive natural language explanations, making the results more accessible to a broader scientific audience or aiding in the formulation of new hypotheses.
    \item \textit{Code generation:} LLMs can assist in the practical aspects of SR by generating code to set up SR experiments, implement the discovered models in simulation environments, or interface SR algorithms with data acquisition systems.
    \item \textit{Bridging data and background knowledge:} LLMs could generate or formalize constraints and search drivers for SR based on background knowledge such as qualitative rules or existing theory. Compact, causal models produced by SR could then be integrated with the help of LLMs into a broader simulation framework. This synergy could potentially bridge the gap between purely data-driven discovery and theory-driven simulation, enabling more powerful and flexible modeling of complex physical phenomena.
\end{itemize}

\section{The Meeting}

The meeting on {\it Symbolic Regression in the Physical Sciences}, held on the 28$^{\rm th}$ and 29$^{\rm th}$ of April, 2025, was a snapshot of the current state of affairs in the field and attempted to cover a broad range of topics. Methodology was at the forefront with a number of talks presenting novel techniques for searching, classifying and, most notably, introducing a level of statistical rigour which has been mostly absent in the field.

Harry Desmond presented the main ideas behind Exhaustive Symbolic Regression and advocated for the use of Descripion Length as a particularly useful approach for ranking expressions, while Roger Guimer\'a explained the strategy behind the Bayesian Machine Scientist, invoking the similarity between Bayesian model selection and Statistical Physics.  Cristina Cornelio advocated for two algorithms -- AI-Descartes and AI-Hilbert -- which incorporate some of the axiomatic and formal proof methods inherent in mathematics, and Geoffrey Bomarito presented the strengths of introducing posterior sampling in genetic programming searches for optimal expressions. Bogdan Burlacu presented a novel approach for duplicate detection using Zobrist hashing while, with similar motivation, Fabricio Olivetti de Franca proposed the use of equality graphs as a compact way of storing expressions which can facilitate the identifying of recurrences. Etienne Russeil presented his approach for using multiple data sets to go from independent experiments to general laws.

There was an opportunity in the meeting to discuss fully developed packages with impressive results: Miles Cranmer presented PySR, a widely used genetic programming based symbolic regression code, J. Nathan Kutz advocated for the use of SHRED, applying sparse regression in the latent space, and William La Cava discussed his multi-objective symbolic regression framework, BRUSH, and   model interpretability in the medical domain.

A key aspect of the workshop was a close look at applications. As well as the examples presented in the previous papers, Deaglan Bartlett showed how analytic emulators for the power spectrum of large scale structure were far more effective than current, neural net based ones. Tariq Yasin showed how ESR could be used to infer analytic profiles directly from weak lensing data from clusters. Evgeniya Kabliman deployed SR to determine analytic expressions for the material properties of metallic alloys, while Steven Abel was able to construct efficient emulators for beyond-standard-model physics. Finally, Andrei Constantin unearthed an intriguing statistical property in the mathematical formulation of physical laws, akin to Zipf's law in the analysis of written texts.

\section{Challenges and Outlook}

% \gk{Should be extended the main topics from the summary of the event: priors to find physically plausible / interpretable equations, uncertainty quantification/propagation, identification of hierarchical models}

{Despite its successes and immense potential, symbolic regression faces ongoing challenges. Scalability to high-dimensional problems remains difficult because the space of possible expressions grows rapidly with the number of inputs. Robustness to noise, outliers, and systematic measurement errors is essential for extracting genuine physical insight rather than reproducing artefacts. Computational limits also constrain the feasibility of searching large expression spaces, motivating the development of more efficient heuristics and search strategies.}

{The risk of identifying mathematically correct but physically meaningless expressions persists. Domain knowledge therefore remains important for formulating well-posed problems, selecting appropriate operators and constraints, and assessing whether discovered relations correspond to genuine mechanisms rather than spurious correlations. Incorporating prior knowledge through existing fits, physical symmetries, or analytically motivated extrapolation behaviour can also reduce the effective search space and improve efficiency compared with purely exploratory approaches.}

{The Royal Society meeting highlighted these issues in detail. Participants emphasised that traditional fit metrics such as mean-squared error are insufficient for assessing the usefulness of a symbolic model, as good interpolation does not guarantee sensible extrapolation or physical plausibility. Additional criteria such as interpretability, simplicity, and robustness under uncertainty were viewed as essential, with approaches like Minimum Description Length providing a principled mechanism for complexity control. There was also sustained discussion about how uncertainty in the data propagates into symbolic expressions, and the need for methods that provide confidence assessments and identify the regimes in which a discovered model can be trusted.}

{The meeting further stressed that symbolic regression’s computational and combinatorial burdens remain fundamental constraints. Even with constrained operator sets or bounded expression complexity, the problem is formally NP-hard, and exhaustive methods face clear trade-offs between optimality, cost, and functional flexibility. Hierarchical or hybrid modelling approaches, in which global structure is shared and dataset-specific parameters vary locally, were proposed as one way to extend the applicability of symbolic regression while mitigating some of these challenges.}

\section{Conclusion}

Symbolic regression stands as a distinctive and increasingly powerful tool in the arsenal of computational science. Its unique ability to derive explicit mathematical models from data endows it with a dual role: it is both a powerful engine for fundamental scientific discovery, particularly in understanding complex emergent phenomena, and a practical instrument for creating robust empirical formulae and efficient emulators. The emphasis on interpretability and the generation of new knowledge sets SR apart from many other machine learning paradigms.

The path forward for symbolic regression in the physical sciences involves continued algorithmic innovation to address challenges of scalability and robustness, deeper integration of domain-specific knowledge to guide the discovery process, creative synergies with complementary AI technologies like large language models, and increasing application to real-world problems in the physical sciences. As these developments unfold, symbolic regression is poised to play an increasingly influential role in accelerating the pace of discovery and deepening our understanding of the complex mathematical tapestry that underlies the physical world.

\bibliographystyle{RS} %%%% .BST file
\bibliography{symreg} %%%%% .Bib file

% \vskip2pc
%
%
%
% \noindent {\bf Please follow the coding for references as shown below.}
%
% \begin{thebibliography}{9}
%
% \bibitem{1} Allwood JM, Cullen JM. 2011 \textit{Sustainable materials:  with both % eyes open}.
% Cambridge, UK: UIT Cambridge. See \href{http://www.withbotheyesopen.com}{http://% www.withbotheyesopen.com}.
%
% \bibitem{2}  MacKay DJC. 2008  \textit{Sustainable energy:  without the hot air}.
%  Cambridge, UK: UIT Cambridge. See \href{http://www.withouthotair.com}{http://% www.withouthotair.com}.
%
% \bibitem{3} Gallman PG. 2011  \textit{Green alternatives and national energy % strategy: the facts
%  behind the headlines}.  Baltimore,\ MD: Johns Hopkins University Press.
%
% \bibitem{4} MacKay DJC. 2013.  Solar energy in the context of energy use, energy % transportation, and
%  energy storage. \textit{Proc. R. Soc. A} {371}.
%
% \end{thebibliography}
%
% \noindent If maintaining .bib file for references, then please use "RS.bst" to % generate the references.
%
% \noindent Example:
%
% \verb+\bibliographystyle{RS}+ %%%% .BST file
%
% \verb+\bibliography{sample}+ %%%%% .Bib file
%
\end{document}